\documentclass[USenglish]{article}

\usepackage[utf8]{inputenc}
\usepackage{natbib}
\usepackage[small]{dgruyter}
\usepackage{microtype}
\setcitestyle{authoryear,open={(},close={)}}
\usepackage{bm}

\begin{document}

% \articletype{Article}

\runningauthor{Nakahara et al.}
\title{Estimating the Effect of Team Hitting Strategies Using Counterfactual Virtual Simulation in Baseball}
  
\author[1]{Hiroshi Nakahara}
\author[2]{Kazuya Takeda}
\author*[3]{Keisuke Fujii}
\affil[1]{Graduate School of Informatics, Nagoya University, Email: nakahara.hiroshi@g.sp.m.is.nagoya-u.ac.jp}
\affil[2]{Graduate School of Informatics, Nagoya University, Email: kazuya.takeda@nagoya-u.jp}
\affil[3]{Graduate School of Informatics, Nagoya University; RIKEN Center for Advanced Intelligence Project; PRESTO, Japan Science and Technology Agency, Email: fujii@i.nagoya-u.ac.jp}
\runningtitle{Estimating the effect of team hitting strategies in Baseball}
\subtitle{}
% abstract: 200 words
\abstract{In baseball, every play on the field is quantitatively evaluated and has an effect on individual and team strategies. 
The weighted on base average (wOBA) is well known as a measure of an batter's hitting contribution.
However, this measure ignores the game situation, such as the runners on base, which coaches and batters are known to consider when employing multiple hitting strategies, yet, the effectiveness of these strategies is unknown. 
This is probably because (1) we cannot obtain the batter's strategy and (2) it is difficult to estimate the effect of the strategies. 
Here, we propose a new method for estimating the effect using counterfactual batting simulation. 
To this end, we propose a deep learning model that transforms batting ability when batting strategy is changed.  
This method can estimate the effects of various strategies, which has been traditionally difficult with actual game data. 
We found that, when the switching cost of batting strategies can be ignored, the use of different strategies increased runs. 
When the switching cost is considered, the conditions for increasing runs were limited.
Our validation results suggest that our simulation could clarify the effect of using multiple batting strategies.}
  \startpage{1}
  \aop

\maketitle
%%%%%%%%%%%%%%%%%%%%%%%%%%%%%%%
\vspace{-0pt}
\section{Introduction}
\vspace{-0 pt}

Advancements in measurement technologies now allow a much higher level of sports analyses.
Especially in baseball, each scene is discrete, making it easy to allocate responsibilities of each play to players.
Many formulas and statistics (stats) were developed and employed to evaluate player's performance (e.g., abilities of batting, pitching, and fielding) from a long time ago \citep{james10, lewis04, click06, tango07, beneventano12, costa12}.
Conventionally, most stats are computed using batting, pitching, and fielding results.
To continually improve measurement, new technologies are being developed that measure sensing data of all Major League Baseball games and make them accessible on data platforms. These data, such as speed and angle of batted balls, allow us to develop stats to estimate players’ performance more accurately.
It enables us to develop stats to estimate player's performance more accurately computed by speed and angle of batted balls, such as expected weighted on-base Average (xwOBA) and expected Earned Run Average (xERA) \citep{xwoba, xera}, 
which can acquire more important information and findings to win.
``Fly ball revolution'', which specifies the effective speed and angle of batted balls needed to hit a long hit.

In \textit{Sabermetrics}, which analyzes baseball based on statistics, 
the importance of each play is defined by the amounts of contributions with a win.
WAR (Wins Above Replacement), which is widely used as an overall player evaluation stat, 
is computed using the number of the contribution of pitching, batting, fielding, and base running.
For batting, it is computed using the weighted on base average (wOBA).
However, wOBA does not consider the game state (e.g., score, out count, runner occupancy, etc.) 
because each player cannot choose when to have a turn at bat or on a pitcher's mound.

In contrast, a player is trained to make his or her play by considering the game state to win the game. 
However, estimating the effect of a strategy with real game data is challenging for two reasons: First, it is difficult to observe or record a team or player's strategy.
Second, various factors affect baseball outcomes (e.g., batting result). 
We thus employ computer simulation to simulate a baseball game and estimate the effect.

Baseball simulations are used for quantifying the effect of strategies.
In this regard, the literature has clarified the optimal batting order for various settings \citep{freeze74, bukiet97, sokol03, hirotsu2016optimal}.
\citet{norman10} conducted a similar study for cricket.
In the analysis of substitution strategies, researchers have employed Markov chain model to simulate games \citep{hirotsu03, hirotsu05modelling}; this approach helps optimize pinch hitting strategies and pitcher substitution strategies. 
Researchers have also focused on evaluating players or plays. 
Some researches focus on evaluating players or plays.
Player's batting ability was quantified by simulating games if they batted in all nine positions in the line-up \citep{cover77}, and the impact of base-running ability was clarified \citep{baumer09}.
Other research \citep{hirotsu2019using} estimated the value of the sacrifice bunt.
Similarly, scholars have estimated the value of the sacrifice bunt \citep{hirotsu2019using} and quantified the effect of rule changes through simulation in various sports \citep{david05, sonne16}.

Following the literature \citep{bukiet97, hirotsu03}, we use a simulator to estimate the effect of strategies using Markov chain modeling. 
A Markov chain is a probabilistic model that describes a series of possible events, where the probability of each event depends only on the state achieved in the previous event. 
This approach is especially suited for baseball. 
Since the transition model for runner advancement used in previous works \citep{bukiet97, hirotsu03} is too simple, we construct a model based on transitions occurring in real games.
However, adopting a different strategy requires a trade-off among stats. 
For example, the probability of long hits decreases when a player prioritizes reaching the base. 
We thus need to estimate a player's ability counter-factually for each strategy. We call this framework ``counter-factual simulation''.

In this study, we consider the effect of batting strategies as one example of using multiple strategies.
% Based on an attitude toward a game context and their own wOBA, batting strategies are categorized to three types: 
Based on the player’s attitude toward the game’s context and the player’s wOBA, we categorize batting strategies into three types:
(1) simply maximizing wOBA without considering other elements, 
(2) considering the game’s context and the player’s wOBA (e.g., light hitting at scoring opportunity), and 
(3) only considering the game context and trying to move a runner to the advanced base (e.g., sacrifice bunt).
Strategy (1) is effective for getting runs as wOBA explains runs well, while some studies show that strategy (3) cannot effectively maximize runs \cite{hirukawa19, tango07}. 
Finally, the effect of strategy (2) has not been clarified because of difficulties in estimating.

In our simulation, we assume that a batter adopts three strategies to maximize runs: (1) normal, (2) prioritizing reaching a base, and (3) prioritizing making a long hit.
Which strategy a player should adopt depends on game contexts.
For estimating the counter-factual ability (prioritizing reaching a base or long hit), we build a ``batting strategy converting'' model that learns the relationships among real players’ batting ability using machine learning. 
This enables us to simulate different batting strategies and, thus, estimate the effect.

The main contributions of this paper are as follows:
(i) estimating the effect of adopting different batting strategies through counter-factual simulations, 
(ii) proposing a new method for a new method for counter-factually converting batting ability using machine learning, and 
(iii) clarifying the relationship between a batting strategy and runs thereof.

\section{Methods}
\label{sec:method}

\subsection{Dataset}

For this study, we used the data of professional Japanese baseball games from 2018 to 2020 provided by the Research Center for Medical and Health Data Science at the Institute of Statistical Mathematics and Data Stadium Inc.
Data acquisition was based on the contract between the baseball league and Data Stadium, Inc., but not between the players and us.
We obtained the data by participating in a competition hosted by the academic organizations. The central idea of this study was independent of the competition (not restricted by the competition).

The data includes pitch-by-pitch data and individual player stats per game.
Pitch-by-pitch data provide, for example, game context (inning, out count, and runner), pitch results, and batting results.
We used pitch-by-pitch data to compute the correspondence between batting result and game context transition.
Individual player stats are indicators of batting performance (plate appearance, hit, homerun, strikeout, and so on) per game, and we use these stats to compute a player's batting ability vector (i.e., the probability of each batting result).

\subsection{Batting simulator}
\label{ssec:batting_simulator}

We use \textit{batting simulator} to simulate a baseball game virtually.
First, we define a batting ability vector $P$ as follows:
\begin{equation}
    P = (p_{1b}, p_{2b}, p_{3b}, p_{hr}, p_{bb}, p_{k}, p_{g}, p_{f}), \label{equation:vector}
\end{equation}
where $p_{1b}, p_{2b}, p_{3b}, p_{hr}, p_{bb}, p_{k}, p_{g}$, and $ p_{f}$ represent the probability of an one-base hit, a two-base hit, a three-base hit, a homerun, a walk and hit by pitch, a strikeout, an easy grounder out, and an easy fly ball out, respectively.
$P$ can be also interpreted as discrete probability distribution, and the sum of each probability is $1$.
Batting results other than those listed above do not occur.
An easy grounder out and easy fly ball out include reaching a base because of the opposing fielder's mistakes; these are thus recorded as a batter's batting result.
We generate nine (the number of batters in a game) $P$, denoted as $\bm{P}$.

\begin{figure}[t]
  \centering
  \includegraphics[width=7cm]{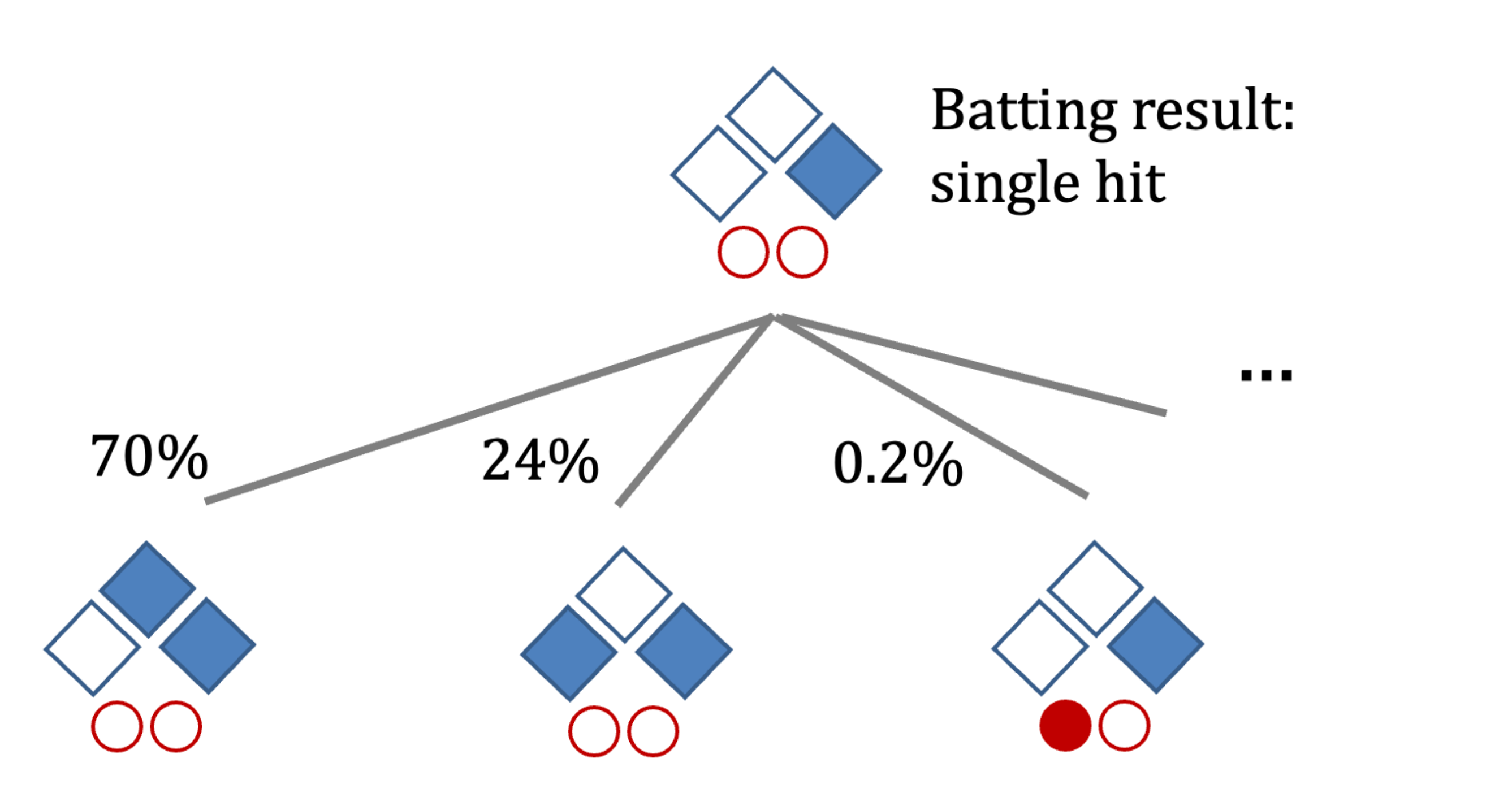}
  \caption{Transition probability when a batter make a single hit where 0-out with 1st-base loaded in \textit{transition function}}
  \label{fig:transition}
\end{figure}

The \textit{Batting simulator} is the function with $\bm{P}$ as the input and team's runs as the output.
Different from real games, it only considers a batting phase (i.e., consists of nine half-innings).
The \textit{Batting simulator} consists of the \textit{batting function} and the \textit{transition function}.
The former is the function with the game context (out count and runner), with $P$ as the input and batting result as the output.
Each batting result is determined by the respective $P$.
A player has three types of $P$ and uses them differently depending on the game context (see \ref{ssec:evaluation}).
The \textit{transition function} is the function with the game context (out count and runner), with batting result as the input, and game context (out count, runner, and run) after batting as the output.
Although extant work \citep{bukiet97, hirotsu03} have mainly used a simple model proposed by \citep{esopo77}, we construct a sophisticated transition model using game context transitions in real games.
Then, using pitch-by-pitch data, we compute the distribution of game context transitions.
Further, the \textit{transition function} represents the function with the previous game context (out count and runner occupation), with batting result as the input and game context (out count, runner occupation, and run) after batting as the output.
For example, when a batter makes a single hit, with $0$ outs with $1$st-base loaded, various transitions are expected as Figure \ref{fig:transition}.
% It works closer to the real ones.
On the contrary, because the \textit{transition function} takes as input only the game context and the batting result, a transition caused by a factor other than the batting result (e.g., wild pitch and stolen base) does not occur. 
This also implies that the running ability of a runner and the fielding ability of a fielder are always average. Despite its simplicity, the \textit{batting simulator} outputs the run distribution similar to real games (see section 3.1). The \textit{batting simulator} repeats the \textit{batting function} and \textit{transition function} for nine innings, and then computes the runs.

\subsection{Counter-factual batting simulation}

\begin{figure}[t]
  \centering
  \includegraphics[width=10cm]{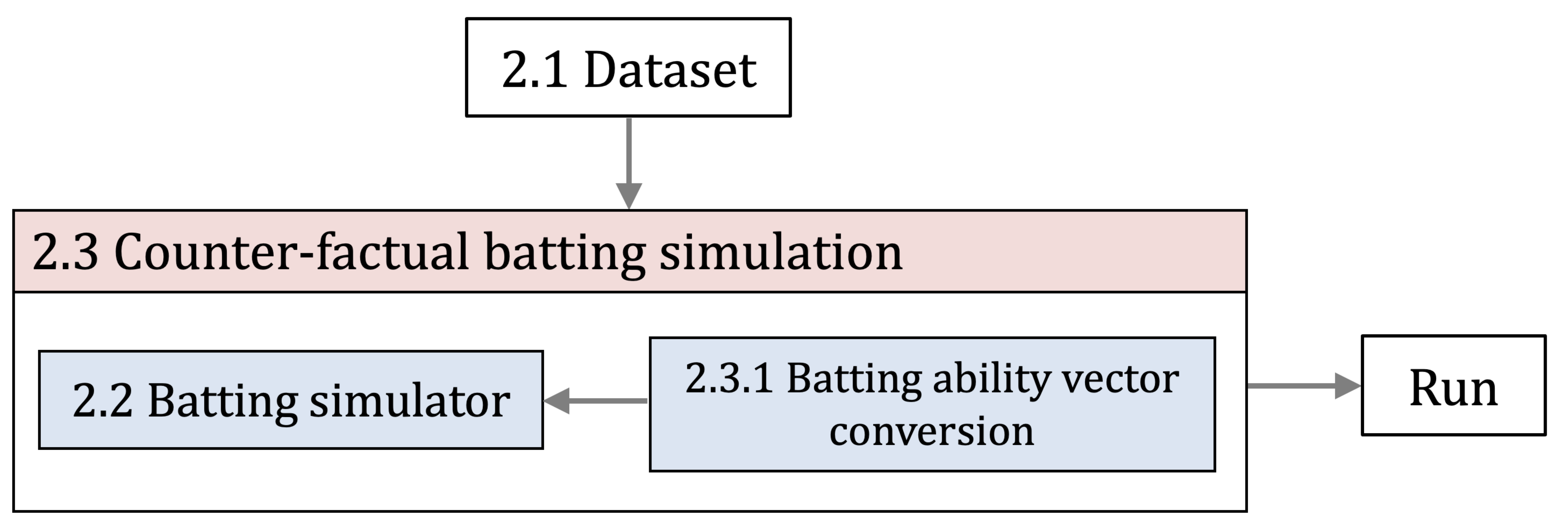}
  \caption{The architecture of \textit{counter-factual batting simulation}}
  \label{fig:entire_concept}
\end{figure}

Next, we propose the \textit{counter-factual batting simulation} using the \textit{batting simulator}, which estimates the effect of the specific strategy with the counter-factual batter's ability conversion (Figure \ref{fig:entire_concept}).
The entire framework of the \textit{counter-factual batting simulation} consists of two phases: (1) generate a counter-factual batter's ability based on their actual performances, and (2) simulate games with the \textit{batting simulator}. 
The architecture of (2) corresponds to the \ref{ssec:batting_simulator}.
The architecture of (2) corresponds to \ref{ssec:batting_simulator}.
Thus, we first describe a general formulation of the problem of converting the batting ability vector, and then describe a formulation for converting the batting ability vector in the case of changing to the on-base/long hit strategy, which is our focus. Finally, we explain how to convert the batting ability vector using machine learning.

\subsubsection{General formulation of the batting ability vector conversion}
\label{ssec:general_formulation}

We consider a change of the individual batting ability vector when converting from the normal strategy $P_n$ to a different strategy (e.g., on-base/long hit prioritized) $P_*$, 
denoted as $\Delta P = P_* - P_{n}$.
This $P_*$ estimation problem is divided into two sub-problems:
(1) estimating a feasible maximum $|\Delta P|$ and 
(2) estimating an appropriate $\Delta P / |\Delta P|$ (unit vector).
We consider it difficult to solve (1), and thus simulate with various $|\Delta P|$.
On the other hand, $\Delta P / |\Delta P|$ can be estimated from the batting ability vectors of real players, and which help us develop a reasonable batting strategy conversion model.

We regard ``batting strategy conversion model'' as the function with $P_n, \Delta \alpha$, and $\Delta wOBA$ as the inputs and $P_*$ as the output, 
where $\Delta \alpha$ and $\Delta wOBA$ are the parameters of the batting strategy conversion model.
$\alpha$ is the magnitude of the batting strategy conversion and can be computed with batting ability vector.
$\Delta \alpha$ is the difference in $\alpha$ before and after conversion and is computed by $\Delta \alpha = \alpha_* - \alpha_{n}$. 
$\Delta wOBA$ is computed by $\Delta wOBA = wOBA_* - wOBA_n (\leq 0)$, and 
$\alpha_*$ and $wOBA_*$ indicate $\alpha$ and $wOBA$ after conversion, respectively.
$\Delta wOBA$ is the cost of the batting strategy conversion 
because we consider that some amount of wOBA reduces when a batting strategy changes.
Because estimating appropriate $\Delta wOBA$ is difficult, 
we simulate with various settings.

\subsubsection{Converting the on-base/long hit prioritized batting ability vector}

We assume that a batter chooses three batting strategies: 
(1) normal, (2) on-base prioritized, and (3) long hit prioritized strategies.
$\alpha$ is defined as follows:
\begin{equation}
     \alpha = f(P) = \frac{\Delta_{1b} \times p_{1b} + \Delta_{bb} \times p_{bb}}{\sum_{\{1b, bb, 2b, 3b, hr\} \in r} \Delta_{r} \times p_r}, \label{equation:alpha}
\end{equation}
where $\Delta_r$ is the run value \footnote{Contribution to runs per a batting result does not consider game context.} of batting result $r$, and 
$\Delta_{1b}=0.437, \Delta{bb}=0.294, \Delta{2b}=0.786, \Delta{3b}=1.117$, and $\Delta{hr}=1.408$ \citep{delta17}.
In other words, $\alpha$ is the percentage of singles and walks in the scoring contribution of the batter with $P$. 

The batting ability vectors for the normal, on-base prioritized, and long hit prioritized strategies denote $P_{n}, P_{o}$, and $P_{l}$ respectively, and the $\alpha$ corresponding to each batting ability vector are $\alpha_{n}, \alpha_{o}$, and $\alpha_{l}$.
We set $P_n, P_o$, and $P_l$ such that $\alpha_{l} < \alpha_{n} < \alpha_{o}$ is satisfied.
In other words, we define a strategy with a larger $\alpha$ than the normal strategy as the on-base prioritized strategy and a strategy with a smaller $\alpha$ as the long hit prioritized strategy. 
For simplicity, all conversions satisfy $\alpha_{o} - \alpha_{n} \simeq \alpha_{n} - \alpha_{l}$ for a player's $\alpha_n, \alpha_o $, and $\alpha_l$. 
Note that the conversion also considers the constraints on $\Delta wOBA$, as described in \ref{ssec:general_formulation}.

\subsubsection{Batting ability vector conversion with machine learning}

Next, to compute the nonexistent batting ability vector, 
we construct the batting strategy conversion model using machine learning, 
which is trained by the relationships among batters who have similar $P$.

\begin{figure}[t]
  \centering
  \includegraphics[width=7cm]{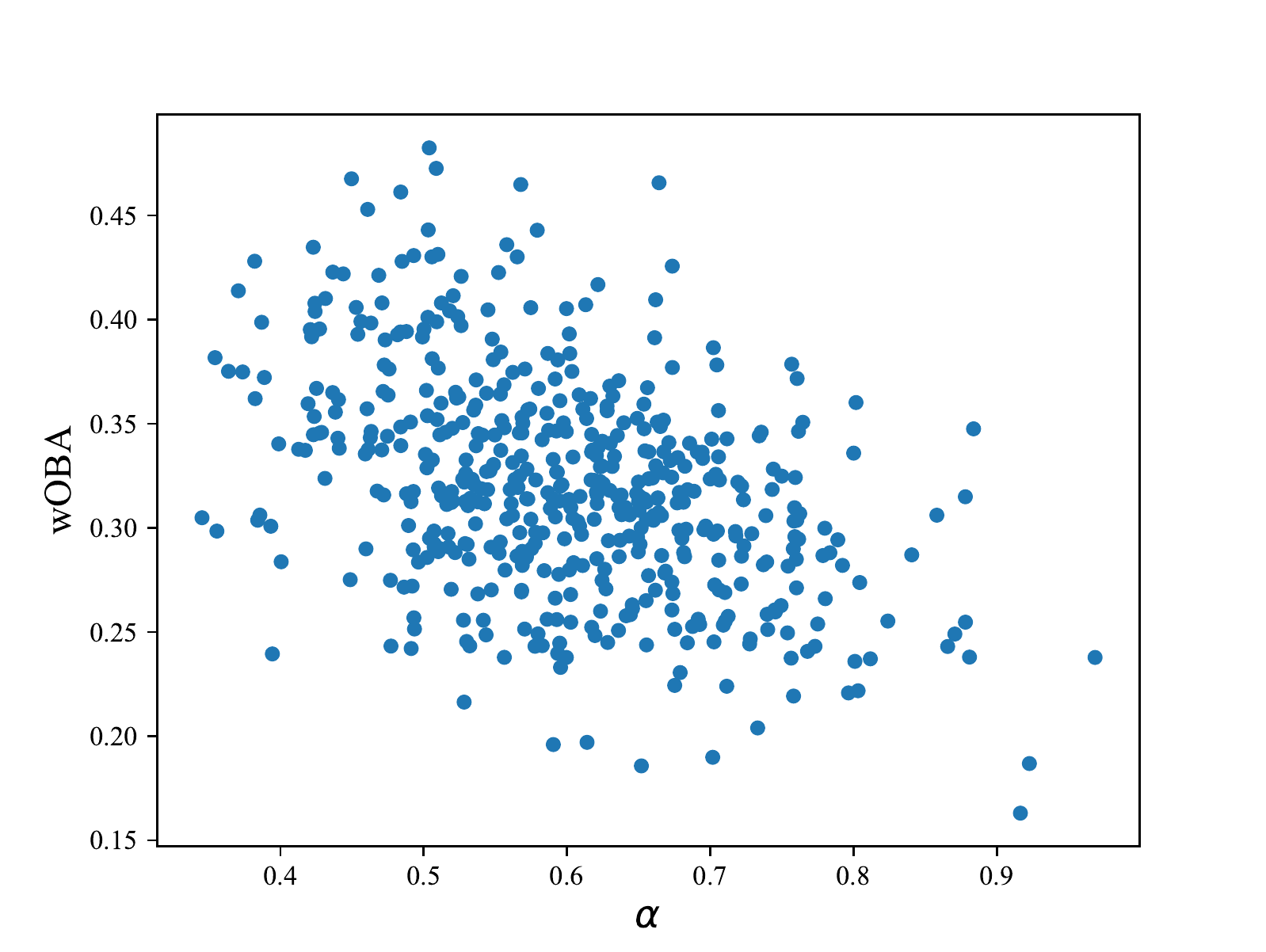}
  \caption{Each batter's $\alpha$ and $wOBA$ with $100+$ at-bat in 2018-2020.}
  \label{fig:map}
\end{figure}

At first, using each batter's batting stat over $100$ at-bat per season, 
we computed $502$ batting ability vectors (Figure \ref{fig:map}).
Then, $P$ were matched for all combinations, which corresponds to $P$ before/after the batting strategy conversion.
Ultimately, $N = {}_{502} C_2 = 126,253$ combinations were generated.
For each pair, $\Delta \alpha$ and $\Delta wOBA$ were computed and used as the train data for the batting strategy conversion model.

We employed a multi-layer perceptron to build the batting strategy conversion model 
because the batting strategy conversion is assumed to be a non-linear conversion.
The input was $P', \Delta \alpha$, and $\Delta wOBA$, 
and the output was $P'_* - P'$, 
where $P'$ denotes that the $8$-dimension vector $P' = (p_{1b}, p_{2b}, p_{3b}, p_{hr}, p_{bb}, p_{k}, p_{g})$.
This is because, based on the constraint that the sum of the probabilities is 1, $p_{f}$ is determined if $P'$ is computed.
The middle layer has $2$ layers with $100$ neurons each and the rectified linear unit (ReLU) was used as the activation function.
The loss function was as follows:
\begin{equation}
\begin{split}
    loss = \frac{1}{N} \sum_{i=1}^N \left\{ 
        \sum_{\bar{p'_i} \in \bar{P}'_i,\hat{p}'_i \in \hat{P}'_i} (\bar{p}'_i - \hat{p}'_i)^2 
        - \alpha \sum_{\hat{p}'_i \in \hat{P}'_i} \min(\hat{p}'_i, 0)\right. \\
        \left. + \beta (wOBA(\bar{P}'_i) - wOBA(\hat{P}'_i))^2,  \right\} \label{equation:loss}
\end{split}
\end{equation}
where $\bar{P'}$ and $wOBA(P')$ represent $P'$ in the train data and wOBA of a player whose batting ability vector is $P'$, respectively.
The first and third terms refer to the mean squared error of the batting ability vector and wOBA, respectively.
The second term is the constraint probability is greater than or equal to $0$.
The soft-max function is not used, because it shows inferior performance to the function employed.
$\alpha$ and $\beta$ represent each constraint's weight; we set $\alpha=0.1$ and $\beta=0.5$ empirically. 
Eighty percent of data was used for training and twenty percent of data were used for validation.

\begin{table}[]
\begin{center}
\caption{Batting stats in the simulation. 
OBP is the on-base average and SLG is the slugging percentage \citep{obp, slg}.}
\begin{tabular}{|c|c|c|c|c|}
\hline
  & OBP& SLG &  wOBA & $\alpha$ \\ \hline
1 & $.337$ & $.377$ & $.320$  & $.64$ \\
2 & $.324$ & $.369$ & $.310$  & $.65$ \\
3 & $.393$ & $.476$ & $.383$  & $.54$ \\
4 & $.360$ & $.464$ & $.363$  & $.50$ \\
5 & $.335$ & $.411$ & $.331$  & $.59$ \\
6 & $.329$ & $.408$ & $.327$  & $.58$ \\
7 & $.316$ & $.369$ & $.307$  & $.58$ \\
8 & $.288$ & $.331$ & $.278$  & $.63$ \\
9 & $.292$ & $.308$ & $.273$  & $.68$ \\ \hline
\end{tabular}
\end{center}
\label{tab:lineup}
\end{table}

\subsection{Evaluation}
\label{ssec:evaluation}

$\bm{P}$ was computed using the average batting ability vector by batting order.
We used all Pacific league games data from 2018 to 2020 (Table \ref{tab:lineup}) given that DH system is adopted in all Pacific league games.
$\Delta \alpha$ and $\Delta wOBA$ were set as $\Delta \alpha=\{0, 0.05, 0.1, 0.15, 0.2, 0.25, 0.3\}, \Delta wOBA=\{0, -0.005, -0.01, -0.015\}$, respectively.
Unless otherwise noted, the condition for employing the on-base prioritized strategy is as follows: 
(1) the out count is zero or 
(2) the second and/or third base is loaded.
Similarly, the conditions for employing the long hit prioritized strategy are as follows:
(1) the out count is two and 
(2) neither the second nor the third base is loaded.
We simulated $100,000$ games for each combination of parameters and obtained the average runs and its standard error.

\section{Results}
\subsection{Model validation}

\begin{figure}[t]
  \centering
  \includegraphics[width=7cm]{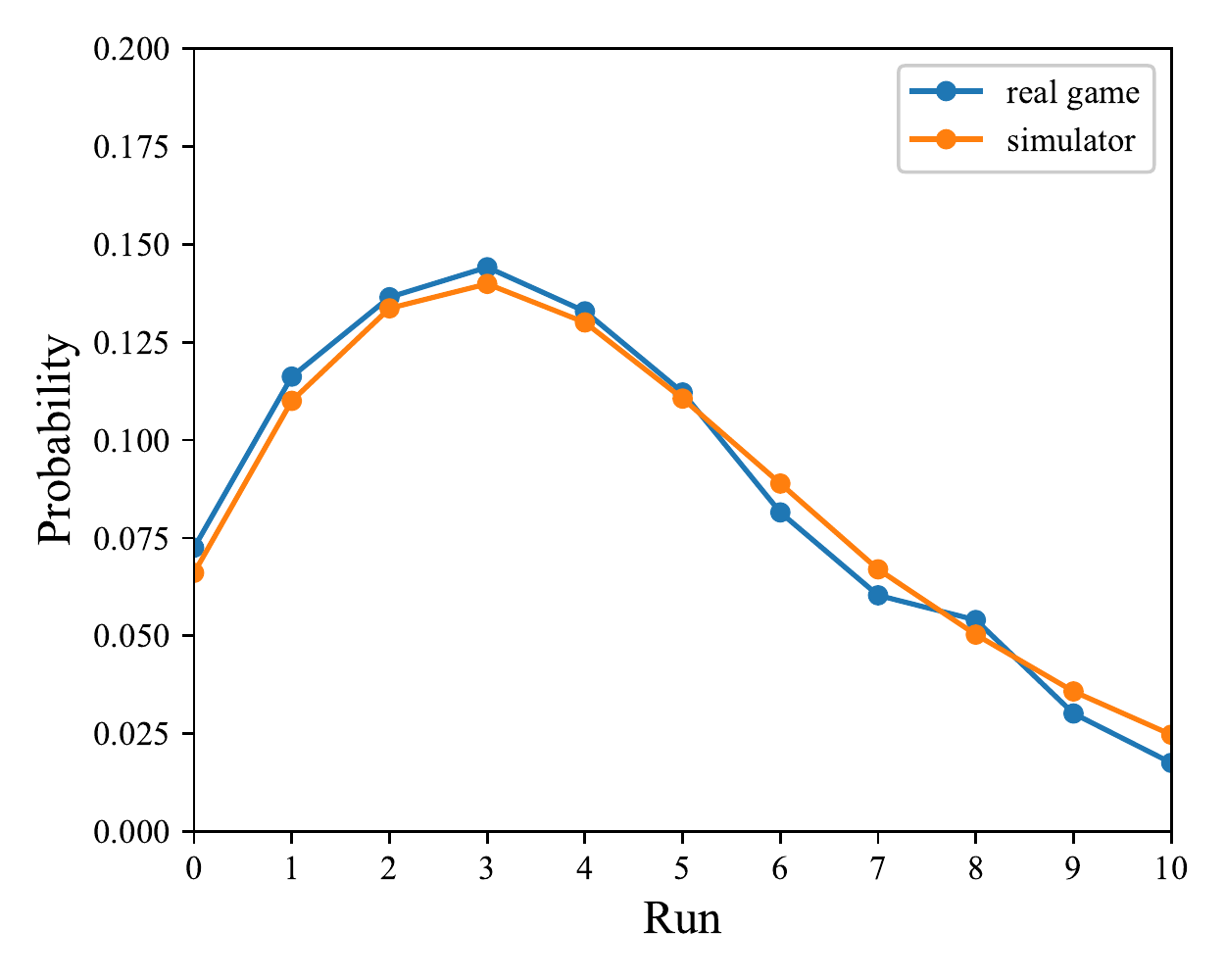}
  \caption{The run distribution comparison between real games and simulation results.}
  \label{fig:run_comparision}
\end{figure}

For the \textit{batting simulator}, the average runs scored in the simulator was $4.27$ per game compared with $4.22$ for all Pacific league games held in 2018--2020.
Figure \ref{fig:run_comparision} shows the run distribution of the simulation and real games.
We considered that the average runs and the run distribution were similar appropriately.
For the \textit{batting strategy conversion model}, mean squared error of $P'$ and wOBA as well as the probability constraint were $1.11 \times 10^{-3}, 4.00 \times 10^{-4}, 0$, respectively. The results suggest that the simulator and the hitting strategy conversion model could output realistic values.

% \subsection{戦略変更の程度、wOBA損失と得点をパラメータとした場合}
\subsection{Effect of using multiple strategies}
\label{ssec:effect_total}

\begin{figure}[t]
  \centering
  \includegraphics[width=7cm]{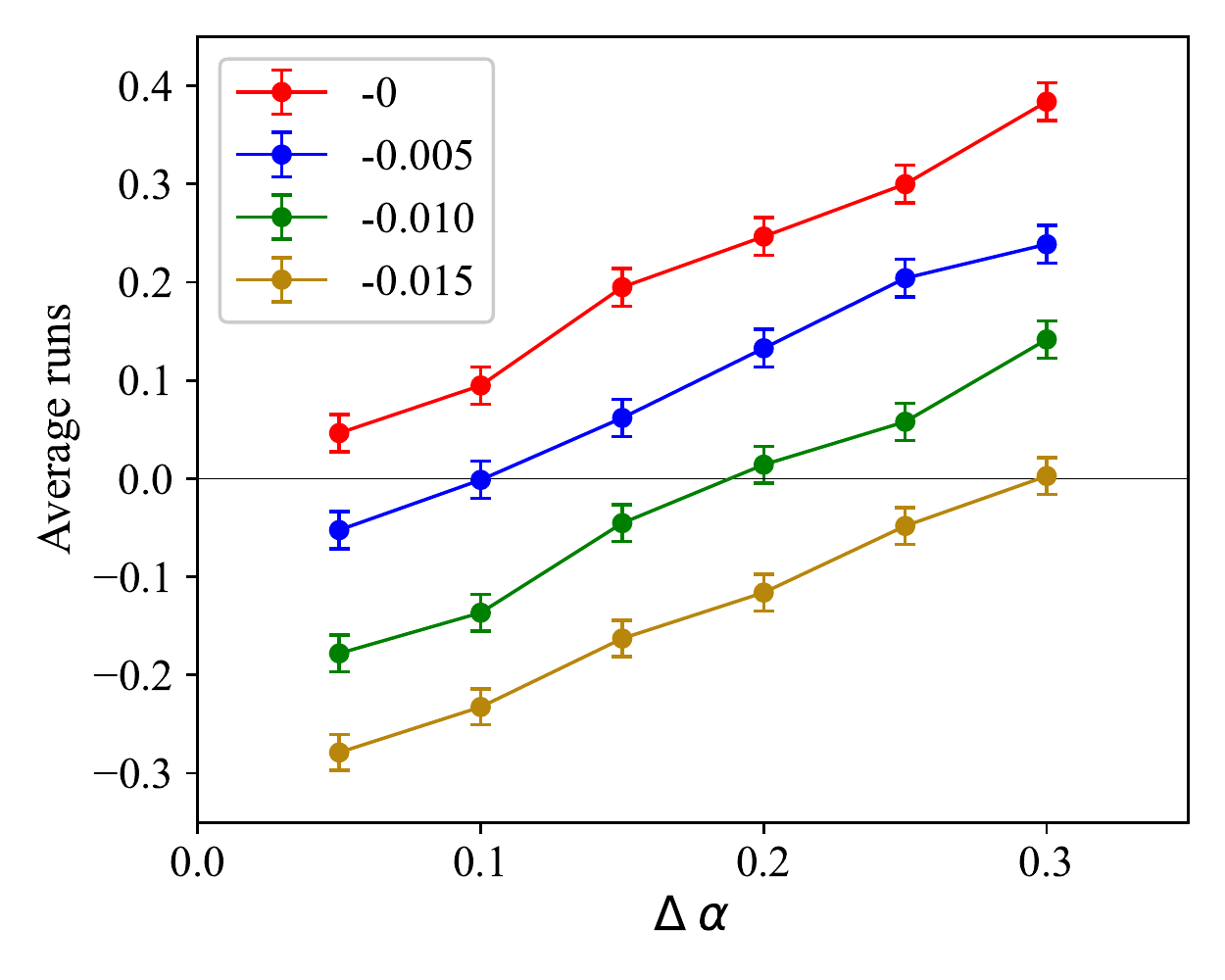}
  \caption{Runs by the combination of $\Delta \alpha$ (X-axis) and $\Delta wOBA$ (plot color).
  The runs with normal strategy are set to the baseline($0.0$).}
  \label{fig:result_a}
\end{figure}

Figure \ref{fig:result_a} shows the result of simulation with  parameters $\Delta \alpha$ and $\Delta wOBA$.
Note that the average runs are expressed as the difference from the baseline ($4.27$), which is fixed with the normal strategy.
When $\Delta wOBA=0$, the average runs increase as $\Delta \alpha$ increased, and was greater than the baseline.
This shows that using multiple batting strategies increases the average runs when the wOBA loss is ignored.
On the other hand, if we consider wOBA loss, the average runs increase, where 
(1) $\Delta \alpha \geq 0.15$ and $\Delta wOBA \geq -0.005$, 
or (2) $\Delta \alpha \geq 0.2$ and $\Delta wOBA \geq -0.01$.

\subsection{Effect by strategy activation conditions}
\label{ssec:effect_trigger}

We also clarified the relationship between strategy activation conditions and average runs.
Although we should perform simulations with all combinations of parameters (out count and runner), it is difficult to simulate all settings ($2^{3 \times 8}$). 
Generally, the on-base prioritized strategy is employed when the run expectancy is high, whereas the long hit prioritized strategy is employed when the run expectancy is low.
Thus, each strategy activation condition was defined by the run expectancy.

More specifically, the on-base prioritized strategy was employed when $RE_c \geq \theta_o$, and long hit prioritized strategy was employed when $RE_c \leq \theta_l$.
Note that $RE_c$ represents the run expectancy of  game context $c$ has, and $\theta_o$ and $\theta_l$ denote the thresholds of the on-base/long hit prioritized strategy.
The on-base prioritized strategy was employed if $RE_c \geq \theta_o$, whereas long hit prioritized strategy was employed if $RE_c \leq \theta_l$.
A normal strategy was employed if both conditions were not satisfied.
The other parameters were fixed as $\Delta \alpha=0.1$ and $\Delta wOBA=-0.005$.

\begin{figure}[t]
  \centering
  \includegraphics[width=8.5cm]{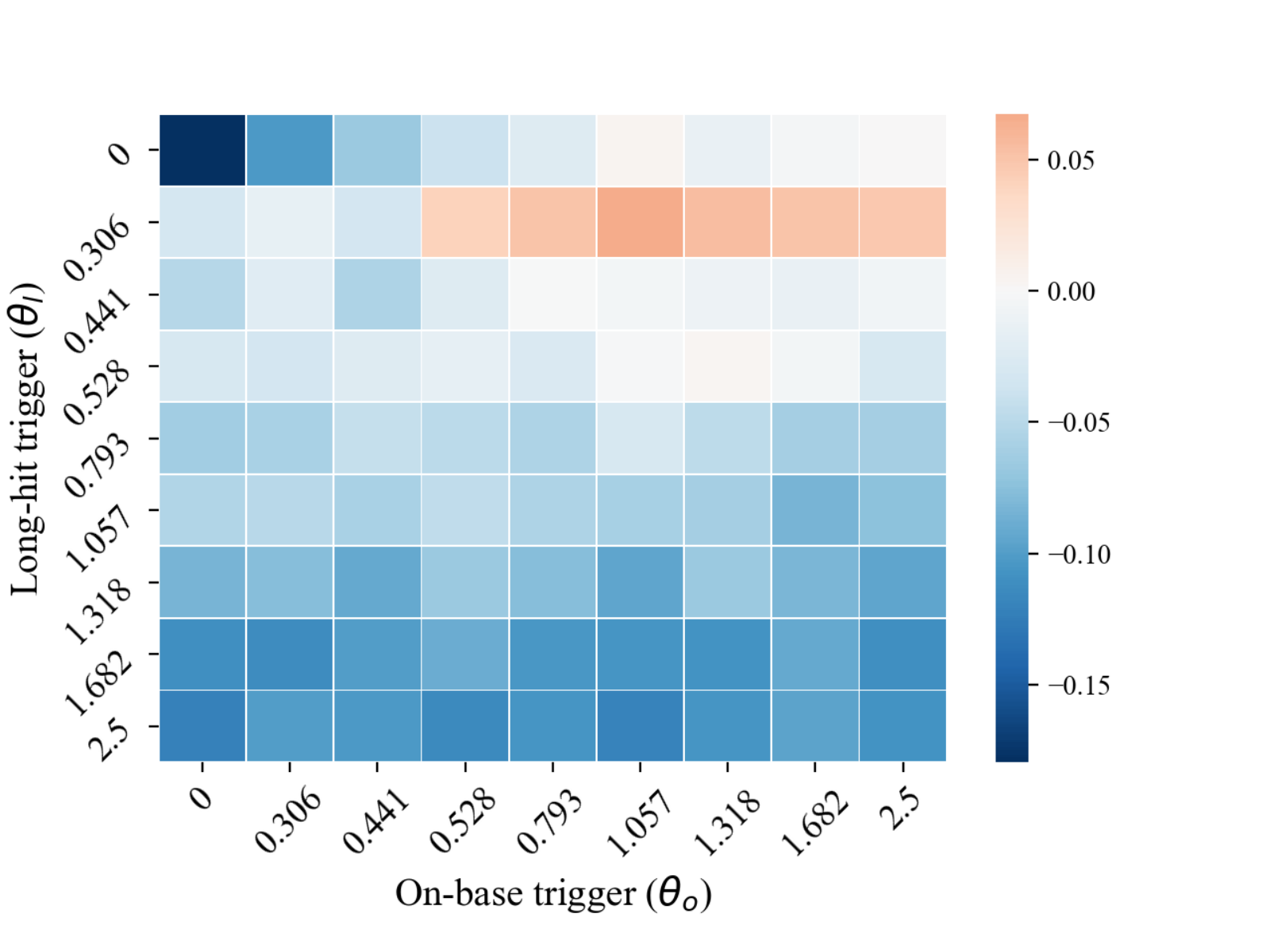}
  \caption{Runs by the combination of $\theta_o$ and $\theta_l$. 
  The X-axis shows the on-base prioritized strategy's trigger.
  The higher the $\theta_o$, the less frequently the on-base prioritized strategy is employed.
  Similarly, the Y-axis shows the long hit prioritized strategy's trigger. The higher the $\theta_l$, the more frequently the long hit prioritized strategy is employed.
The normal strategy was employed if both conditions were not satisfied.
The other parameters were fixed as $\Delta \alpha=0.1$ and $\Delta wOBA=-0.005$.
  The runs with normal strategy were set to the baseline($0.0$) as Figure \ref{fig:result_a}.}
  \label{fig:result_trigger}
\end{figure}

Figure \ref{fig:result_trigger} shows the result.
The lower left corner shows looser triggering conditions (employed frequently) and the upper right corner shows stricter triggering conditions (employed less frequently).

The average runs increased when the strategy activation conditions were strict (employed less frequently).
The combinations of $\theta_o$ and $\theta_l$ with the highest runs ($4.33$) were $\theta_o = 1.057$ and $\theta_l = 0.306$.
$\theta_o = 1.057$ corresponds to $0$ out with second base loaded, and $\theta_l = 0.306$ corresponds to $2$ out with first base loaded.

\section{Discussion} %  and Conclusion
In this study, we proposed the \textit{counter-factual batting simulation} and the \textit{batting strategy conversion model} to simulate a game in which the players use multiple batting strategies.
The validation results suggest that our simulation would be useful to clarify the effect of using multiple batting strategies.

As shown in section \ref{ssec:effect_total}, we clarified that using multiple batting strategies increased the average runs when the wOBA loss was ignored.
Thus, the appropriate batting ability vector changes depending on the situation.
The strategy of changing batting ability vectors depending on the situation was thus confirmed to be reasonable as a batter does.

More specifically, multiple batting strategies ($\Delta \alpha=0.2, \Delta wOBA=-0.005$) increased $0.13$ runs per game compared with the normal strategy.
Assuming that each team has $143$ games per a season and $10$ runs worth $1$ win, 
However, WAR does not evaluate team batting, as it does not consider game context in the evaluation. 
Given that WAR quantifies the contribution to runs (wins), we must evaluate the contributions of team hitting.

On the other hand, when we did consider the wOBA loss, we found that the conditions for increasing runs were limited.
Thus, it is better not to use multiple batting strategies if $\Delta wOBA \leq -0.015$.
For a $P$ of average ability, the standard deviation of the wOBA distribution per $400$ at-bat is $0.026$, whereas the value of $0.015$ is comparatively lower. 
Although changing batting strategies is welcomed in Japan, the cost of doing so for a batter may be too small. 
Players and coaches should be careful when employing multiple batting strategies given the trade-offs thereof.

As shown in \ref{fig:result_trigger}, the average runs increased where strategy activation conditions were strict (lower employed).
In most situations, the runs were lower than in the normal strategy.
Runs were thus highly dependent on $\Delta \alpha$ and $\Delta wOBA$, and ideally, a simulation is done for each setting in practical.
 
The question ``Should the player use multiple batting strategies?'' cannot be answered from these results, because the relationship between $\Delta \alpha$ and $\Delta wOBA$ remains unclear, and 
it is difficult to estimate this relationship using real game data.
For future studies, we suggest experiments to estimate this relationship.

Further, the batting strategy conversion model still has room for improvement.
Note that it does not consider the batter's character except for $\alpha$, and the conversion might be inappropriate.
For example, we can consider the ratio of $BB\%$ to $1B\%$, where $BB\%$ and $1B\%$ are the percentage of walk and first-base hit, respectively.
Then, player A's ability with high $BB\% / 1B\%$ may be converted into an ability with low $BB\% / 1B\%$.
In other words, only $\Delta \alpha$ is considered, and other facts are ignored.
Clustering players before building the model may reasonably address this problem.

\section*{Acknowledgments}
This work was supported by JST PRESTO (JPMJPR20CA).

%\bibliographystyle{DeGruyter}
% \bibliography{main}

\begin{thebibliography}{25}
  \providecommand{\natexlab}[1]{#1}
  \providecommand{\url}[1]{\texttt{#1}}
  \expandafter\ifx\csname urlstyle\endcsname\relax
    \providecommand{\doi}[1]{doi: #1}\else
    \providecommand{\doi}{doi: \begingroup \urlstyle{rm}\Url}\fi
  
  \bibitem[Baumer(2009)]{baumer09}
  B.~S. Baumer.
  \newblock Using simulation to estimate the impact of baserunning ability in
    baseball.
  \newblock \emph{Journal of Quantitative Analysis in Sports}, 5\penalty0 (2),
    2009.
  
  \bibitem[Beneventano et~al.(2012)Beneventano, Berger, and
    Weinberg]{beneventano12}
  P.~Beneventano, P.~D. Berger, and B.~D. Weinberg.
  \newblock Predicting run production and run prevention in baseball: the impact
    of sabermetrics.
  \newblock \emph{Int J Bus Humanit Technol}, 2\penalty0 (4):\penalty0 67--75,
    2012.
  
  \bibitem[Bukiet et~al.(1997)Bukiet, Harold, and Palacios]{bukiet97}
  B.~Bukiet, E.~R. Harold, and J.~L. Palacios.
  \newblock A markov chain approach to baseball.
  \newblock \emph{Operations Research}, 45\penalty0 (1):\penalty0 14--23, 1997.
  
  \bibitem[Click and Keri(2006)]{click06}
  J.~Click and J.~Keri.
  \newblock \emph{Baseball between the numbers: Why everything you know about the
    game is wrong}.
  \newblock Perseus Books Group, 2006.
  
  \bibitem[Costa et~al.(2012)Costa, Huber, and Saccoman]{costa12}
  G.~B. Costa, M.~R. Huber, and J.~T. Saccoman.
  \newblock \emph{Reasoning with Sabermetrics: Applying Statistical Science to
    Baseball's Tough Questions}.
  \newblock McFarland, 2012.
  
  \bibitem[Cover and Keilers(1977)]{cover77}
  T.~M. Cover and C.~W. Keilers.
  \newblock An offensive earned-run average for baseball.
  \newblock \emph{Operations Research}, 25\penalty0 (5):\penalty0 729--740, 1977.
  
  \bibitem[D'ESOPO(1977)]{esopo77}
  D.~D'ESOPO.
  \newblock The distribution of runs in the game of baseball.
  \newblock \emph{Optimal Strategies in Sports, Ladany}, 1977.
  
  \bibitem[Forrest et~al.(2005)Forrest, Beaumont, Goddard, and Simmons]{david05}
  D.~Forrest, J.~Beaumont, J.~Goddard, and R.~Simmons.
  \newblock Home advantage and the debate about competitive balance in
    professional sports leagues.
  \newblock \emph{Journal of Sports Sciences}, 23\penalty0 (4):\penalty0
    439--445, 2005.
  
  \bibitem[Freeze(1974)]{freeze74}
  R.~A. Freeze.
  \newblock An analysis of baseball batting order by monte carlo simulation.
  \newblock \emph{Operations Research}, 22\penalty0 (4):\penalty0 728--735, 1974.
  
  \bibitem[Hirotsu and Bickel(2019)]{hirotsu2019using}
  N.~Hirotsu and J.~E. Bickel.
  \newblock Using a markov decision process to model the value of the sacrifice
    bunt.
  \newblock \emph{Journal of Quantitative Analysis in Sports}, 15\penalty0
    (4):\penalty0 327--344, 2019.
  
  \bibitem[Hirotsu and Eric~Bickel(2016)]{hirotsu2016optimal}
  N.~Hirotsu and J.~Eric~Bickel.
  \newblock Optimal batting orders in run-limit-rule baseball: a markov chain
    approach.
  \newblock \emph{IMA Journal of Management Mathematics}, 27\penalty0
    (2):\penalty0 297--313, 2016.
  
  \bibitem[Hirotsu and Wright(2003)]{hirotsu03}
  N.~Hirotsu and M.~Wright.
  \newblock A markov chain approach to optimal pinch hitting strategies in a
    designated hitter rule baseball game.
  \newblock \emph{Journal of the Operations Research Society of Japan},
    46\penalty0 (3):\penalty0 353--371, 2003.
  
  \bibitem[Hirotsu and Wright(2005)]{hirotsu05modelling}
  N.~Hirotsu and M.~Wright.
  \newblock Modelling a baseball game to optimise pitcher substitution strategies
    incorporating handedness of players.
  \newblock \emph{IMA Journal of Management Mathematics}, 16\penalty0
    (2):\penalty0 179--194, 2005.
  
  \bibitem[Hirukawa(2019)]{hirukawa19}
  K.~Hirukawa.
  \newblock \emph{Introduction to Sabermetrics}.
  \newblock Suiyosha, 2019.
  
  \bibitem[James(2010)]{james10}
  B.~James.
  \newblock \emph{The new Bill James historical baseball abstract}.
  \newblock Simon and Schuster, 2010.
  
  \bibitem[Lewis(2004)]{lewis04}
  M.~Lewis.
  \newblock \emph{Moneyball: The art of winning an unfair game}.
  \newblock WW Norton \& Company, 2004.
  
  \bibitem[MLB.com(2015{\natexlab{a}})]{xera}
  MLB.com.
  \newblock Expected era (xera).
  \newblock \url{https://www.mlb.com/glossary/statcast/expected-era},
    2015{\natexlab{a}}.
  \newblock Accessed: 2021-07-05.
  
  \bibitem[MLB.com(2015{\natexlab{b}})]{xwoba}
  MLB.com.
  \newblock Expected weighted on-base average (xwoba).
  \newblock \url{https://www.mlb.com/glossary/statcast/expected-woba},
    2015{\natexlab{b}}.
  \newblock Accessed: 2021-07-05.
  
  \bibitem[MLB.com(2022{\natexlab{a}})]{obp}
  MLB.com.
  \newblock On-base average (obp).
  \newblock \url{https://www.mlb.com/glossary/standard-stats/on-base-percentage},
    2022{\natexlab{a}}.
  
  \bibitem[MLB.com(2022{\natexlab{b}})]{slg}
  MLB.com.
  \newblock Slugging percentage (slg).
  \newblock
    \url{https://www.mlb.com/glossary/standard-stats/slugging-percentage},
    2022{\natexlab{b}}.
  
  \bibitem[Norman and Clarke(2010)]{norman10}
  J.~M. Norman and S.~R. Clarke.
  \newblock Optimal batting orders in cricket.
  \newblock \emph{Journal of the Operational Research Society}, 61\penalty0
    (6):\penalty0 980--986, 2010.
  
  \bibitem[Okada(2017)]{delta17}
  Y.~Okada.
  \newblock \emph{Delta Baseball Report1}.
  \newblock Suiyosha, 2017.
  
  \bibitem[Sokol(2003)]{sokol03}
  J.~S. Sokol.
  \newblock A robust heuristic for batting order optimization under uncertainty.
  \newblock \emph{Journal of Heuristics}, 9\penalty0 (4):\penalty0 353--370,
    2003.
  
  \bibitem[Sonne and Keir(2016)]{sonne16}
  M.~W. Sonne and P.~J. Keir.
  \newblock Major league baseball pace-of-play rules and their influence on
    predicted muscle fatigue during simulated baseball games.
  \newblock \emph{Journal of sports sciences}, 34\penalty0 (21):\penalty0
    2054--2062, 2016.
  
  \bibitem[Tango et~al.(2007)Tango, Lichtman, and Dolphin]{tango07}
  T.~M. Tango, M.~G. Lichtman, and A.~E. Dolphin.
  \newblock \emph{The book: Playing the percentages in baseball}.
  \newblock Potomac Books, Inc., 2007.
  
  \end{thebibliography}

\end{document}